\documentclass[runningheads]{llncs}
\usepackage{graphicx}

\usepackage{tikz}
\usepackage{comment}
\usepackage{amsmath,amssymb} 
\usepackage{color}
\usepackage[nosort, nocompress]{cite}
\usepackage{booktabs}
\usepackage{multirow}
\usepackage{arydshln}
\usepackage{subfig}
\usepackage{wrapfig}
\def\red#1{\textcolor[rgb]{1,0,0}{#1}}
\def\blue#1{\textcolor[rgb]{0,0,1}{#1}}

\newcommand{\doublecheck}[1]{\textcolor{black}{#1}}
\newcommand{\keypoint}[1]{\vspace{0.1cm}\noindent\textbf{#1}\quad}
\newcommand{\etal}{\textit{et al}. }

\newcommand{\cut}[1]{}
\usepackage[accsupp]{axessibility}  

\begin{document}
\pagestyle{headings}
\mainmatter
\def\ECCVSubNumber{1366}  

 \title{Adaptive Fine-Grained Sketch-Based Image Retrieval}
\authorrunning{Bhunia \etal} 
\author{Ayan Kumar Bhunia\textsuperscript{1} \hspace{.1cm}  Aneeshan Sain\textsuperscript{1,2} \hspace{.1cm} Parth Hiren Shah\thanks{Interned with SketchX}  \\ Animesh Gupta\textsuperscript{$\star$} \hspace{.1cm}  Pinaki Nath Chowdhury\textsuperscript{1,2} \hspace{.1cm} Tao Xiang\textsuperscript{1,2} \hspace{.1cm} Yi-Zhe Song\textsuperscript{1,2}}
\institute{\textsuperscript{1}SketchX, CVSSP, University of Surrey, United Kingdom.  \\
\textsuperscript{2}iFlyTek-Surrey Joint Research Centre on Artificial Intelligence.\\
{\tt\small \{a.bhunia, a.sain, p.chowdhury, t.xiang, y.song\}@surrey.ac.uk}}

\maketitle
\vspace{-0.5cm}
\begin{abstract}
The recent focus on Fine-Grained Sketch-Based Image Retrieval (FG-SBIR) has shifted towards generalising a model to new categories without any training data from them. In real-world applications, however, a trained FG-SBIR model is often applied to both new categories and different human sketchers, i.e., different drawing styles. Although this complicates the generalisation problem, fortunately, a handful of examples are typically available, enabling the model to adapt to the new category/style. In this paper, we offer a novel perspective -- instead of asking for a model that generalises, we advocate for one that quickly adapts, with just very few samples during testing (in a few-shot manner). To solve this new problem, we introduce a novel model-agnostic meta-learning (MAML) based framework with several key modifications: (1) As a retrieval task with a margin-based contrastive loss, we simplify the MAML training in the inner loop to make it more stable and tractable. (2)  The margin in our contrastive loss is also meta-learned with the rest of the model. (3) Three additional regularisation losses are introduced in the outer loop, to make the meta-learned FG-SBIR model more effective for category/style adaptation. Extensive experiments on public datasets suggest a large gain over generalisation and zero-shot based approaches, and a few strong few-shot baselines.


\vspace{-0.2cm}
\keywords{FG-SBIR, Meta-Learning, Category and Style Adaptation.}
\end{abstract}

\vspace{-0.9cm}
\section{Introduction}
\vspace{-0.2cm}

Significant progress has been made towards making sketch an input modality for image retrieval \cite{bhunia2020sketch, wang2015sketch, SBIR_imbalance, zhang2018generative,liu2017deep, sampaio2020sketchformer, collomosse2019livesketch, shen2018zero}. As an input modality complementary to text, sketch finds its competitive advantage especially when it comes to fine-grained instance-level retrieval~\cite{bhunia2020sketch, pang2020solving, BMVC_hierarchy, pang2017cross}, where the problem lies with intra-category retrieval as opposed to the conventional category-level setting \cite{collomosse2017sketching, collomosse2019livesketch}.  

\begin{figure}[]
\centering
        \includegraphics[ width=0.48\linewidth]{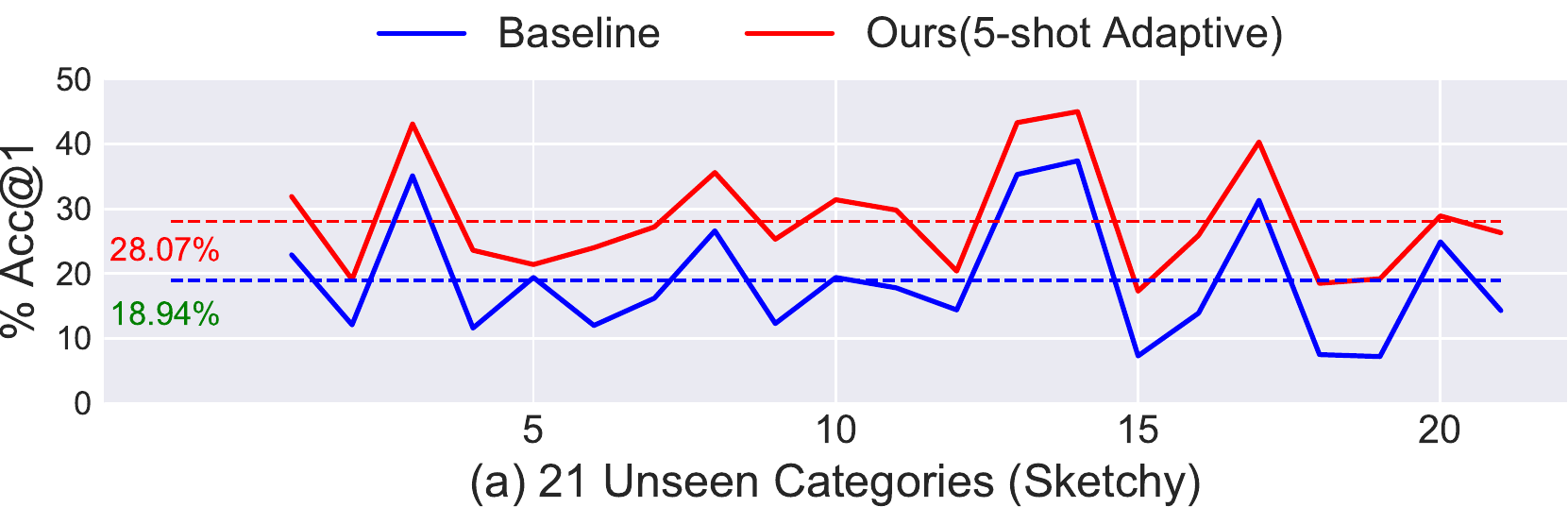} 
        \includegraphics[ width=0.48\linewidth]{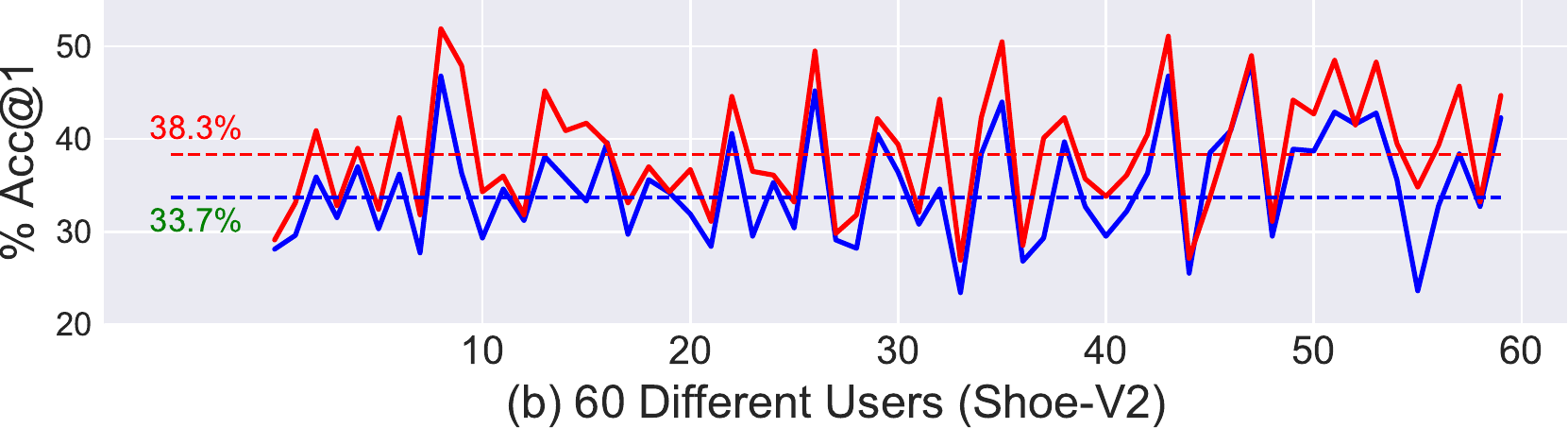}
        \includegraphics[ width=1.0\linewidth]{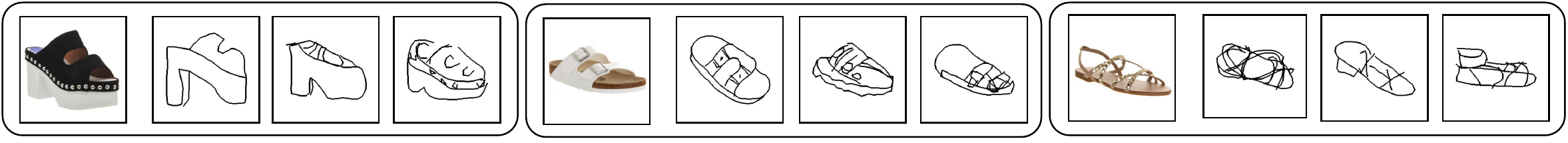}
\vspace{-0.75cm}
  \caption{Graphs illustrating  how (a) \emph{category-adaptive} FG-SBIR (b) \emph{user-adaptive} FG-SBIR  can significantly improve the retrieval performance on unseen categories and users using only 5 samples during inference, respectively. (c) Examples showing the varying \emph{style} of sketching with different level of abstraction for the same photo with respect to different users (drawers).  
}
\label{fig:Fig1}
\vspace{-0.6cm}
\end{figure}

Early attempts at fine-grained sketch-based image retrieval (FG-SBIR) mainly focused on tackling the sketch-photo domain gap, where triplet-based networks have by now been established as the de facto choice \cite{QianIJCV2020, bhunia2020sketch, pang2020solving, song2017deep}. As performance under the supervised learning setting have recently started to saturate, the research focus has shifted onto the problem of data scarcity, where the challenge is to build generalisable and zero-shot models for unseen categories \cite{dey2019doodle, pang2019generalising, dutta2019semantically}. However, retrieval performances of these models are typically much weaker compared to supervised models. We attribute this to two factors (i) the stringent assumption of no additional sketch-photo pairs from the new categories, and more importantly, (ii) drawing styles of input sketches vary significantly amongst different users (see Figure \ref{fig:Fig1}-c) -- the latter of which remains untackled to date.
\vspace{-.5cm}

The first contribution is thus a \cut{\emph{new} and} \emph{practical} problem setting, namely category/style adaptive FG-SBIR. Instead of asking for a model that \textit{generalises} to categories \cite{dey2019doodle, pang2019generalising, dutta2019semantically}, we advocate for one that \textit{quickly adapts}.  That is, we are after a \textit{single} FG-SBIR model that can \textit{quickly} adapt to a new style/category, with just a few samples \textit{during testing}. Achieving this offers a best-of-both-worlds solution -- (i) the model has a better chance at adaptation having observed new style/category data, as opposed to no data for generalisation or zero-shot, and (ii) the few samples requirement still falls within the practical remit of sketch data, i.e., one can always sketch just a few. We show by experiments that our quick adaptation \doublecheck{(few-shot)} approach (category-level) offers about $6\%$ \cut{performance} gain over generalisation-based models ($9-10\%$ over no adaptation baseline), with just 5 new samples \textit{during testing only} (Figure \ref{fig:Fig1} offers a summary).  Our ultimate vision for commercial adaption of FG-SBIR is therefore -- to deploy a single model, where the end users can easily adapt to their specific categories and drawing styles, by sketching very few ($\le$10) new samples. 

Our second contribution is to devise a novel meta-learning framework to solve this new problem of adaptive FG-SBIR. Essentially, we build upon model-agnostic meta-learning (MAML) \cite{finn2017MAML}, which learns a common initialisation point encoding knowledge shared across different tasks such that it adapts quickly for a new task (i.e., specific category or user) using a few training samples. Unique to conventional few-shot approaches \cite{oreshkin2018tadam, snell2017prototypical}, this suits us ideally as it yields \textit{one} model, and needs only a few (usually one) gradient update steps during testing. 

However, getting a MAML-based framework to work with the specific problem of FG-SBIR is non-trivial.  First,  unlike few-shot classification -- for which MAML was initially proposed -- triplet-based (margin-based contrastive loss) cross-modal FG-SBIR networks typically involve three forward passes for anchor, positive and negative instances; and adopting MAML off-the-shelf would additionally incur heavy computation due to their second-order gradient computation \cite{raghu2019rapid} during backpropagation. We propose to side-step these difficulties by performing inner loop updates only for the final joint-feature embedding layer (see Figure~2).  This importantly avoids an over-fitted model during adaptation, as not all parameters are updated during adaptation process. Besides performing meta-learning upon intermediate latent-space, \cut{meta-learning an intermediate latent space produced by a feature embedding network} we also \emph{meta-learn the margin} used in the contrastive loss to adapt it to new categories. To further tackle the sketch-photo domain gap \cite{ganin2015unsupervised}, we additionally introduce a domain discrimination module to regularize the intermediate latent-space at the outer loop. 

Our next contribution lies with how to tailor our meta-learning framework to best work with user- and category-wise adaptation. For that, we aim to make the intermediate latent space, upon which meta-learning is performed, be category/style discriminative. This discriminative objective is handled through an auxiliary classification head for category-level adaptation. Whereas, due to absence of abundant data for every user, we substitute for an auxiliary contrastive learning head for the style-adaptation setting. Furthermore, we add an extra semantic reconstruction head to encourage category-level transfer (akin to zero-shot SBIR \cite{dey2019doodle, dutta2019semantically, dutta2020semantically}). In brief, both category and style adaptation are regularised by domain adaptation and the discriminative objective, while semantic relatedness is specifically modelled for category-level adaptation.  

Our contributions can be summarised as follows: (a) We set out a vision for practical FG-SBIR by proposing a new problem setting, where rather than seeking for generalisation, we advocate for quick adaptation at testing time. (b) We introduce a novel FG-SBIR framework based on gradient-based meta-learning that adapts to a new category or user sketching style based on a few training examples. (c) The framework is based on the existing MAML but with significantly different formulations  tailored for the specific challenges of either category or style-level adaptation. (d) Extensive experiments on public datasets suggest a significant increase in performance over generalisation and zero-shot approaches, and few strong few-shot baselines.

\vspace{-0.7cm}
\begin{figure}
\begin{wrapfigure}{l}{0.6\linewidth}
\vspace{-1cm}
\includegraphics[width=\linewidth]{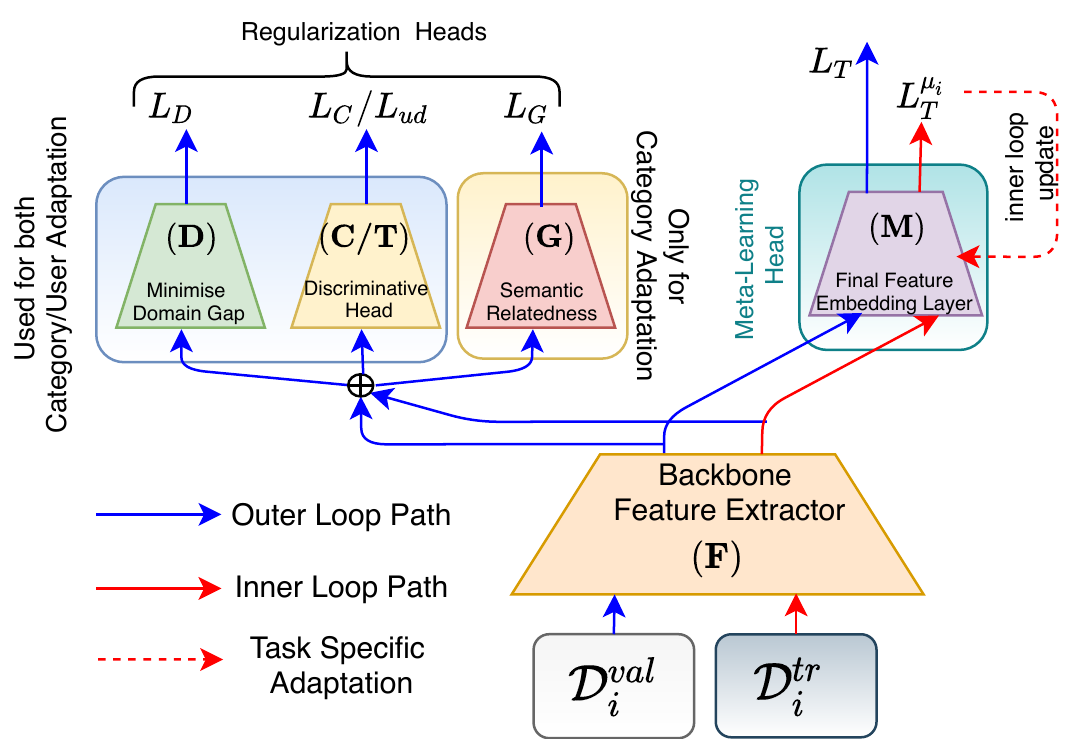}
\vspace{-1.2cm}
\end{wrapfigure}
Fig. 2: Our framework for category/user adaptation involves a bi-level optimisation process. While the inner loop (red) aims to adapt via pseudo-updating $\mathbf{M}$ using support set, the outer loop (blue) executes meta-optimisation to learn better initialisation parameter. Moreover, meta-learning is performed in the intermediate latent space $\mathbf{F(\cdot)}$ which is regularised by auxiliary heads. Discriminative head is modelled by either $\mathbf{C}$ (with loss $L_C$) or $\mathbf{T}$ (with loss $L_{ud}$) for category and user level adaptation, respectively. $\mathbf{G}$ is only used for category-level adaptation. More details is in Section \ref{meta_fgbsir}. Note that \emph{only the inner loop path} (red) is used to obtain category/user-specialised model during inference. 
\label{fig:Fig2}
\vspace{-1cm} 
\end{figure}
\setcounter{figure}{2}



\vspace{-0.2cm}
\section{Related Works}
\vspace{-0.2cm}
\noindent \textbf{Fine-Grained SBIR:} Unlike category-level SBIR \cite{collomosse2017sketching, liu2019semantic, dutta2020semantically, zhang2018generative, collomosse2019livesketch, sain2021stylemeup, Sketch3T, bhunia2022diy, bhunia2021vectorization}, the goal of fine-grained SBIR \cite{pinaki2022PartialSBIR, bhunia2022subset, bhunia2021more, chowdhury2022fs}  is \emph{instance-level} matching. While Li \etal \cite{li2014fine} first introduced it using deformable part models and graph matching, Yu \etal \cite{yu2016sketch} employed deep learning via deep triplet network to learn a common embedding space from heterogeneous domains. Later methods improved upon this via attention mechanism with higher order retrieval loss \cite{song2017deep}, \cut{text tags \cite{song2017fine}, cross-domain image generation \cite{pang2017cross}, cross-modal hierarchical co-attention \cite{BMVC_hierarchy},} reinforcement learning for on-the-fly retrieval \cite{bhunia2020sketch}  and using mixed modal jigsaw solving for a better pre-training strategy \cite{pang2020solving}. In this paper, we introduce a new fine-grained SBIR setting, i.e. category/style adaptive FG-SBIR, which is relevant to real-world applications given the performance gain.

\noindent \textbf{Cross-Category Generalisation for SBIR:}  Mostly studies have been done on category level SBIR \cite{dutta2019semantically, dey2019doodle, liu2019semantic, dutta2020semantically} for cross-category generalisation. Starting with sketch-photo translation pseudo tasks using conditional GANs \cite{yelamarthi2018zero} that learn embedding, zero-shot SBIR has been handled by regularising embedding space, with semantic information across different classes by reconstructing word-vectors \cite{dey2019doodle}, semantically paired cycle consistency \cite{liu2019semantic}. Cross-category generalisation for FGSBIR has only been attempted via \cut{which adopted} a domain-generalisation approach by modelling a universal manifold of prototypical visual sketch traits \cite{pang2019generalising} to dynamically represent the sketch/photo. Contrary to the  domain-invariant representation learning \cite{pang2019generalising}, we follow a few-shot adaptation \cite{finn2017MAML} approach and additionally address the user-specific style-adaptation problem. 

\noindent \textbf{Meta-Learning:} 
Meta-learning has been studied intensively for quick model adaptation to new tasks with few training samples. Representative methods such as memory network \cite{oreshkin2018tadam} or metric-based \cite{snell2017prototypical} meta-learning methods are mostly architecture dependant \cite{choi2020adaptiveScene} and generally designed for few-shot classification. They are thus unsuitable for our retrieval model adaptation problem.  Recently there has been a significant attention towards optimisation based meta-learning algorithms \cite{finn2017MAML, finn2018probabilistic, li2017metaSGD} due to their model agnostic nature. In particular,  model-agnostic meta-learning (MAML) \cite{finn2017MAML} aims to learn optimal initialisation parameters that allows quick adaptation at test-time using few gradient descent updates. Later on, it was further augmented with sets of tricks to stabilise the training in MAML++ \cite{antoniou2018trainMAML}, learnable learning rate in MetaSGD  \cite{li2017metaSGD},  meta-optimisation in a low-dimensional latent space in LEO \cite{rusu2019LEO}, or a simplified inner-loop update by recently introduced Sign-MAML~\cite{fan2021sign}.  While MAML in theory can be applied to our problem, we have to introduce a number of significant modifications to make it more tractable and well suited to the specific challenges associated with category/style-adaptive FG-SBIR.   

To eliminate disparity among users \cite{lane2011activity}, several user adaptive models \cite{hsieh2015facial, bhunia2021metahtr} are developed for activity and emotion recognition by transfer learning \cite{garcia2020adaptiveEmotion}, and user interface via recurrent network \cite{soh2017UI}. Conversely, we use meta-learning that realises our goal of a user adaptive AI agent. 


\vspace{-0.4cm}
\section{Methodology}
\vspace{-0.4cm}
\keypoint{Overview:} We devise a fine-grained SBIR framework that could be instantly adapted for either specific category or user. The category/user specific training and testing data consists of $\mathcal{D}^\mathrm{S} = \{\mathcal{D}_1^\mathrm{S}, \mathcal{D}_2^\mathrm{S}, \cdots, \mathcal{D}_{|\mathrm{N^S}|}^S \ni |\mathrm{N^S}| > 1 \}$ and $\mathrm{\mathcal{D}^T = \{\mathcal{D}_1^T, \mathcal{D}_2^T, \cdots, \mathcal{D}_{|N^T|}^T \ni |N^T| \geq 1 \}}$, where $\mathrm{N^S}$ and $\mathrm{N^T}$ are the disjoint sets of training and testing categories/users (styles) respectively, i.e. $N^S \cap  N^T = \emptyset$. Furthermore, we have access to smaller fine-tuning sets of data corresponding to every testing category/user as $\mathcal{D}^\mathrm{F} = \{\mathcal{D}_1^\mathrm{F}, \mathcal{D}_2^\mathrm{F}, \cdots, \mathcal{D}_{|N^T|}^\mathrm{F} \ni |N^T| \geq 1 \}$ for category/user specific instantaneous adaptation. The i-th category/user in either of the three sets consists of $\mathrm{K^i}$ paired sketch ($\mathrm{x}$) and photo ($\mathrm{y}$) images as $\mathrm{\mathcal{D}_i = \{x_j, y_j\}_{j=1}^{K^i}}$. During training, we intend to meta-learn a retrieval model $F_M$ with the optimal initialisation point, by modelling the shared knowledge across different category/user from the training set $\mathcal{D}^{S}$ -- such that it can quickly adapt to any new category/user using few examples. During inference,  given $\mathrm{K}$ sketch-photo pairs from the fine-tuning set of a particular category/user $\mathcal{D}_i^F$,  we obtain a category/user \emph{specialised model} via a single gradient update:  $F_M \mapsto F_M^i$.

\vspace{-0.3cm}
\subsection{Baseline FG-SBIR Model} \label{basemodel}
\vspace{-0.1cm}
We use Siamese network with spatial attention \cite{dey2019doodle} as the baseline retrieval model. It consists of two  components: (i) Given a photo or a rasterized sketch image $I$, we extract backbone feature map $\mathrm{B = f_{B}(I)} \in \mathbb{R}^{h\times w \times c}$ where $f_{B}$ is  initialised from a pre-trained InceptionV3 \cite{bhunia2020sketch} model; and  h, w, and c represent the height, width and channels respectively. The attention normalised feature are fused with backbone feature via a residual connection to give  $\mathrm{B_{att} = B + B\cdot f_{att}(B)}$, followed by a global-average pooling operation to get a latent feature vector representation of size $\mathbb{R}^{c}$. This CNN comprising $f_{B}$ and $f_{att}$  produces an intermediate latent feature embedding $\mathbf{F}$, parameterised by $\mathrm{\mathbf{\theta_{F}}}$. (ii)  The extracted feature vector is passed through a fully-connected layer  followed by $l_2$ normalisation  to embed the photo and sketch images into a shared embedding space of dimension $\mathbb{R}^d$. We call this component as $\mathbf{M}$ with parameters $\mathrm{\mathbf{\theta_{M}}}$. 

Overall, the final representation is obtained through staged operation denoted as $\mathbf{M \circ F}$. The training data are triplets $\{a, p, n\}$ containing sketch anchor, positive and negative photos respectively. Accordingly, \textit{Triplet loss} is used for training \cite{weinberger2009metric_learn_margin}, which aims at increasing the distance between sketch anchor and negative photo $\beta^{-} = \left \| \mathbf{M \circ F}(a) -  \mathbf{M \circ F}(n) \right \|_{2}$ while reducing that between sketch anchor and positive photo $\beta^{+} = \left \|  \mathbf{M \circ F}(a) -  \mathbf{M \circ F}(p) \right \|_{2}$. Let $\mu$ be the margin-hyperparameter, the  triplet loss calculated across a batch of size N: 

\vspace{-0.4cm}
\begin{equation}
\label{triplet_loss}
  L_T =  \frac{1}{N}\sum_{i=1}^{N} max \{0, \mu + \beta^{+}_{i} - \beta^{-}_{i}\}.
\vspace{-0.2cm}
\end{equation}

\noindent In multi-category FG-SBIR where a single model handles instance-specific retrieval from multiple categories (e.g. Sketchy \cite{sangkloy2016sketchy}), hard triplets are used in training, i.e., the negative photo is from the same class but of different instances.  

\vspace{-0.3cm}
\subsection{Background: Gradient based Meta-Learning} \label{maml}
\vspace{-0.2cm}
Given a set of related tasks with some distribution $p(\mathcal{T})$, the objective of MAML \cite{finn2017MAML} is to meta-learn a good initialisation $\theta$ of some parametric model $f_{\theta}$ such that only a few examples  is necessary to adapt to any new task $\mathcal{T}_{i} \sim p(\mathcal{T})$. Each sampled task $\mathcal{T}_{i} \in \{D^{tr}, D^{val}\}$ consists of a support set of examples $D^{tr}$ and a query/target set of examples $D^{val}$.  Given  $\mathcal{D}^{tr}$ and a loss function $\mathcal{L}_{\mathcal{T}_{i}}$, the parameters $\theta$ are first adapted to $\theta_{i}^{'}$ using one or more gradient descent updates via \emph{inner-loop} (performs adaptation) feedback as follows: $\theta_{i}^{'} \leftarrow \theta - \alpha \nabla_{\theta}\mathcal{L}_{\mathcal{T}_{i}}^{\mathcal{D}^{tr}}(f_{\theta})$. 
The inner-loop learning rate $\alpha$ can either be fixed, or meta-learned concurrently like Meta-SGD \cite{li2017metaSGD}. 
As the aim is to optimise $\theta$ in such a way that one or a few gradient based updates will enable maximally effective performance on any $\mathcal{T}_{i}$,  meta-optimisation is conducted in the \emph{outer-loop} update, with respect to $\theta$ as: $\theta \leftarrow \theta - \beta \nabla_{\theta} \sum_{\mathcal{T}_{i} \sim p(\mathcal{T})} \mathcal{L}_{\mathcal{T}_{i}}^{\mathcal{D}^{val}}(f_{\theta_{i}^{'}})$. Updating $\theta$ via the outer-loop essentially implies \emph{gradient through a gradient}, or differentiating through the inner-loop to minimise the meta-objective using task-specific adapted models $f_{\theta_{i}^{'}}$ on their corresponding \emph{target set} $\mathcal{D}^{val}$. This is computationally demanding due to the second order gradient computation on $\theta$.

\vspace{-0.3cm}
\subsection{Meta-Learning for FG-SBIR}\label{meta_fgbsir}
\vspace{-0.2cm}
\keypoint{Overview:} A number of modifications are needed to make MAML suitable for our category/user adaptive FG-SBIR problem.  The first challenge of adopting MAML for FG-SBIR is to alleviate the high computational cost brought about by the nested optimisation in the inner and outer loops. Inspired by a recent study \cite{raghu2019rapid} which suggests that inner loop simplification in MAML has little impact on its effectiveness, we   exclude  $\mathbf{F}$ from the inner loop update, and only meta-learn the final feature embedding layer $\mathbf{M}$ by adapting parameter $\mathrm{\mathbf{\theta_{M}}}$ inside the inner loop. In other words, meta-learning is performed on the intermediate latent feature space extracted by  $\mathbf{F}$. 

We further introduce regularizers for handling problems specific to fine-grained SBIR. More specifically, there is a meta-learning head of final feature embedding layer $\mathbf{M}$  (used during inner loop update) and multiple regularisation heads upon the extracted latent representation $\mathbf{F}(\cdot) \in \mathbb{R}^c$ (see Figure 2). These regularizers serve two major purposes: (a) minimising the sketch-photo domain gap in the intermediate latent space, and (b) allowing the intermediate latent space to be more discriminative across different categories/styles. Furthermore, we add an \emph{extra} regularisation head to aid in semantic transfer to unseen categories in category-level adaptation. Note that this is only for category adaptation and \emph{not} used for user style adaptation as no varying semantic concepts exist across different users \cite{dutta2019semantically}. 

Moreover, all regularizers are removed during inference, and we only use $\mathbf{M \circ F}$ where the final joint-feature embedding head $\mathbf{M}$ is updated through a single gradient update using few support set examples for quick adaptation. In a nutshell, during adaptation, $\mathbf{M}$ grabs the specialised knowledge from support set examples to generalise better for a specific target category/user. Next we describe these regularizers in detail.

\keypoint{Minimising Sketch-Photo Domain Gap:} Bridging the domain-gap between sketch and photo images is a key objective  behind learning feature representation in a common embedding space for any sketch-based image retrieval system \cite{dey2019doodle}. Therefore, we add a discriminator $\mathbf{D}: \mathbb{R}^c \mapsto [0,1]$ with parameter $\mathbf{\theta_{D}}$ which learns to predict the domain of an input (i.e. sketch vs photo) from the latent features of size $\mathbb{R}^c$ from $\mathbf{F}$. By maximising this discriminator loss through Gradient Reversal Layer (GRL) \cite{ganin2015unsupervised}, the network $\mathbf{F}$ learns to extract domain-agnostic latent feature upon which the meta-learning head $\mathbf{M}$ can generalise better. Given the binary domain label $t$ (which is 0 and 1 for sketch and photo domain respectively) for input I, the binary cross-entropy loss to train the domain-adaptation head is defined as: 
\vspace{-0.25cm}
\begin{equation}
\mathrm{L_D = t \cdot \log (\mathbf{D}(\mathbf{F}(I))) + (1 - t) \cdot \log (1 - \mathbf{D}(\mathbf{F}(I)))}
\vspace{-0.25cm}
\end{equation}

\keypoint{Discriminative Intermediate Latent Space:}  While triplet loss over the output of $\mathbf{M}$ distinguishes between instances of a particular category, a class discrimination objective \cite{horiguchi2019significance} helps towards learning to separate between different categories in a multi-category FG-SBIR model.  In order to make the latent space $\mathbf{F(\cdot})$ class-discriminative, we add a cross-entropy loss using a classification head $\mathbf{C}: \mathbb{R}^c \mapsto \mathbb{R}^{\mathrm{|N^S|}}$ with parameters $\mathbf{\theta_{C}}$. Let the class label be $\mathrm{\mathbf{c_l} \in N^S}$ with respect to input either sketch or photo image $I$, the classification loss is defined as:
\vspace{-0.15cm}
\begin{equation}
\mathrm{L_C = \texttt{Cross\_Entropy}(\mathbf{c_l}, \texttt{softmax}(\mathbf{C}(\mathbf{F}(I))))}.
\vspace{-0.2cm}
\end{equation}
For some datasets such as QMUL-ShoeV2 \cite{yu2016sketch, bhunia2020sketch} where user-level adaptation is required but all sketch-photo images belong to the \emph{same} shoe category, this classification loss is clearly not applicable. In this case,  we want the intermediate latent space $\mathbf{F}(\cdot)$ to be discriminative across different user's sketching styles instead. As the number of samples per user is limited, we use a metric learning based approach. Concretely, a triplet loss is used where anchor ($a'$) and positive ($p'$) sketch-photo pairs come from the same user, and negative ($n'$) sample is from any other user. Directly imposing triplet loss over the latent space $\mathbf{F(\cdot)}$ can be a very hard constraint \cite{dou2019domain}, potentially hurting generalisation during instant adaptation to new users; thus we use an auxiliary embedding network $\mathbf{T}: \mathbb{R}^{c+c} \mapsto \mathbb{R}^{d'}$, where concatenated features of paired sketch-photo are fed as input. Note that we are imposing triplet loss on the output of $\mathbf{T}$ but the gradient flows back through $\mathbf{F}$ making its $\mathbf{F}(\cdot)$ user-discriminative. Given $\beta'^{-} = \left \| \mathbf{T}(\mathbf{F}(a'))  -  \mathbf{T}(\mathbf{F}(n')) \right \|_{2}$, $\beta'^{+} = \left \| \mathbf{T}(\mathbf{F}(a'))  -  \mathbf{T}(\mathbf{F}(p')) \right \|_{2}$, and $\mu'$ being the margin-hyperparameter, the triplet loss is computed as: 
 

\vspace{-0.3cm}
\begin{equation}
\mathrm{L_{ud} = max\{0, \beta'^{+}  -  \beta'^{-} + \mu'\}}.
\vspace{-0.2cm}
\end{equation}

\keypoint{Semantic Transfer for Category-Level Adaptation:} Unlike single-category FG-SBIR \cite{yu2016sketch, bhunia2020sketch}, multi-category FG-SBIR (e.g. Sketchy) further dictates the transfer of class-specific semantic concept  \cite{fu2015zero} from seen to unseen categories. For that, meta-learning is additionally performed in the semantically enriched intermediate latent-space $\mathbf{F}(\cdot)$ by using relationship between different categories.
Specifically, we use a semantic decoder head over $\mathbf{F}(\cdot)$ to reconstruct the word-embedding representation of the category label with respect to either sketch or photo. Let $\mathbf{G}$ be the semantic decoder (three fully connected layers with ReLU) with parameter $\mathbf{\theta_G}$,  input (sketch or photo) be $I$ with class label $\mathrm{\mathbf{c_l} \in N^S}$ and word-embedding from pre-trained FastText \cite{bojanowski2017enriching} model be $\mathbf{S_w}$= $ \texttt{embedding}(\mathbf{c_l})$. We use simple cosine-similarity distance as semantic reconstruction loss: 

\vspace{-0.3cm}
\begin{equation}
\mathrm{L_S = \frac{1}{2}\bigg(1 - \frac{\langle \mathbf{G}(\mathbf{F}(I)), \; S_w\rangle}{\left \| \mathbf{G}(\mathbf{F}(I)) \right \|_{2} \cdot\left \| S_w\right \|_{2} }\bigg)}
\vspace{-0.1cm}
\end{equation}

\keypoint{Task Sampling:} 
In meta-learning \cite{tim2020metaSurvey}, a model is trained episodically, such that a task sampled in each episode, imitates the few-training-sample scenario appearing during testing.
For us, sampling a task $\mathcal{T}_{i} \sim p(\mathcal{T})$ across a category/user means: (i) we first randomly select the i-th category/user $\mathcal{D}_i^S$ out of $\mathrm{N^S}$ sets of training category/user. (ii) Next, from $\mathcal{D}_i^S$, we construct the  support $\mathcal{D}^{tr}$ and validation set $\mathcal{D}^{val}$ by randomly sampling $\mathrm{K}$ sketch-photo pairs for each respectively. Inner loop is updated over $\mathcal{D}^{tr}$, and outer loop over $\mathcal{D}^{val}$. Within every set, hard negatives are created by selecting different photo instances.

\keypoint{Meta Optimisation on the Loss Margin:} The \emph{triplet loss} contains the hyper-parameter, margin $\mu$, whose optimal value is empirically found to be varying ($\S$ \ref{ablation}) across different categories. Since the intra-class distribution or spread among sampled sketches is unlikely to be identical for each class, it is intuitive to have a class-specific optimal margin value. Therefore, we decide \emph{learning to learn} the margin-hyperparameter inside our meta-learning process that would adaptively decide the optimal $\mu$ value for a specific category at test time. 

For the i-th task, given K-shot training examples from $\mathcal{D}^{tr}_{i} = \{(x_k, y_k)\; | \; k = 1, \cdots, K\}$, the latent representation of sketch ($x_k$) is concatenated with that of its corresponding photo ($y_k$), to obtain the per-instance sketch-photo representation: $f_{xy}^k = \texttt{concat}(\mathbf{F}(x_k), \mathbf{F}(y_k)) \in \mathbb{R}^{c + c}.$ Given the set of all $\mathrm{\{f_{xy}^k\}_{k=1}^{K}}$, different per-instance sketch-photo representations $\{f_{xy}^m, f_{xy}^n\}$ where $(m \neq n)$ are concatenated pair-wise, resulting in a total of $K' = K(K-1)$ pairs. All such pairs can be aggregated into a matrix for task $i$ as $S_i \in \mathbb{R}^{K' \times (2c+2c)}.$ 
This is subsequently processed by a \emph{relational network} $\mathbf{R}$ with parameter $\mathbf{\theta_R}$ that feeds every row-vector to each time step of bidirectional GRU to model the relation among all samples in support set. A max-pool operation is performed over the output from all time steps. The resultant vector is then fed to a linear layer that finally predicts a sigmoid normalised scalar value representing the learnable margin value for each task $i$ as $\mu_i = \mathbf{R}(S_i)$. Therefore, given the task specific triplet loss $L_{T_i}^{ \mu_i}$ (Eqn. \ref{triplet_loss}) with its margin hyperparameter $\mu_i$ and  learning rate $\alpha$ both being meta-learned concurrently following Meta-SGD \cite{li2017metaSGD}, the parameter of the $\mathbf{M}$ is now adapted in the  inner loop using $\mathcal{D}^{tr}_{i}$ as follows: 
\vspace{-0.2cm} 
\begin{equation}\label{innerloop}
\mathbf{\theta'_M} =  \mathbf{\theta_M} - \mathbf{\alpha} \cdot \nabla_{\theta_M} L_{T}^{\mu_i}(\mathbf{\theta_F}, \mathbf{\theta_R}, \mathbf{\theta_M}; \mathcal{D}^{tr}_{i}). \vspace{-0.15cm}    
\end{equation}
 
\noindent On the other side, the overall regularisation loss for \emph{category level adaptation} becomes $L_{reg} = \sum_{a,p,n}\frac{1}{3}(L_D + L_C + L_S)$ which is calculated over anchor (a), positive (p) and negative (n) samples. Similarly, for \emph{user-specific adaptation} the regularisation loss becomes $L_{reg} = \sum_{a,p,n}\frac{1}{3}L_D + L_{ud}$.  As there is no inner-loop step, we calculate the regularisation loss over concatenated samples from both support and validation set together (say $\mathcal{D}_i$). 

\noindent Let all parameters related to regularisation be denoted as $\mathbf{\theta_{reg}}$, e.g., for category adaptation $\mathbf{\theta_{reg}} = \{\mathbf{\theta_{D}}, \mathbf{\theta_{C}}, \mathbf{\theta_{S}}\}$ and user style adaptation $\mathbf{\theta_{reg}} = \{\mathbf{\theta_{D}}, \mathbf{\theta_{T}}\}$. Meta-learning pipeline is trained along with regularisation loss to optimise a combined loss. The optimisation objective for the outer loop is thus formulated as:  

\vspace{-0.4cm} 
 \begin{equation}\label{meta-opt}
  \operatorname*{argmin}_{\mathbf{\theta_F}, \mathbf{\theta_M}, \mathbf{\theta_R}, \alpha, \mathbf{\theta_{reg}}}  L_{T}( \mathbf{\theta_F}, \mathbf{\theta_R}, \mathbf{\alpha}, \mathbf{\theta'_M}; \mathcal{D}^{val}_{i}) + \lambda \cdot L_{reg}( \mathbf{\theta_F}, \mathbf{\theta_{reg}}; \mathcal{D}_i) 
 \end{equation}
 
\vspace{-0.2cm}
\noindent where $\lambda$ is a weighting hyperparameter. Note that the task specific adapted $\mathbf{\theta'_{M}}$ is used to compute a validation loss.
As $\mathbf{\theta'_{M}}$  is dependant on $\mathbf{\theta_{M}}$, $\mathbf{\theta_{R}}$ and $\mathbf{\alpha}$ via \cut{through the} inner-loop update (Eqn. \ref{innerloop}), a higher order gradient is computed in the outer loop optimisation.  Note that the model is updated by averaging gradient over meta-batch size of $\mathrm{B}$ sampled tasks and trained in an end-to-end manner.

\keypoint{\doublecheck{Discussion}:} (a) \emph{Significance of Semantic-Relatedness Loss:} For multi-category FG-SBIR (Sketchy), sketch-photo pairs are from the same category, grouped together using the class discriminative objective. Every category holds a semantic concept, which may help control the positioning of class-specific groups in the embedding space, such that class-specific concepts can be transferred from seen to unseen categories (akin to zero-shot SBIR \cite{dey2019doodle, dutta2019semantically}). This entire objective is handled by the semantic relatedness module. It should not be confused with the instance-specific separation criteria for fine-grained retrieval (\textit{already handled} by triplet-loss). 
Please see $\S$ \textbf{Supp.} for an illustration of latent space. 
(b) \emph{Difference with classical few-shot learning:} Standard few-shot literature usually deals with classification \cite{finn2017MAML}, whereas ours is the \emph{first work employing few-shot adaptation for fine-grained retrieval}. We show potential under two objectives: category and user's style adaptation. (c) \emph{Novelty behind Triplet-Loss+MAML:} Dou \etal \cite{dou2019domain} adopted MAML for \emph{domain generalisation} purpose where triplet loss acts as an auxiliary loss to encourage class specific feature clustering. On the contrary, ours involves a few-shot \emph{adaptation paradigm} which requires executing inner loop update using triplet-loss during inference. Therefore, the design of inner-loop update using triplet loss is more critical to our framework. Furthermore, unlike \cite{dou2019domain}, margin value of inner-loop triplet loss is meta-learned to facilitate better and stable adaptation.  (d) \emph{Why FG-SBIR undergoes such generalization issue (unlike person ID/re-ID):} Domain gap existing across various categories in multi-category FG-SBIR, is much larger than different person-identities in Re-ID, as shape morphology varies highly across new categories (not limited to just human shapes). Note that FG-SBIR model tries to learn \emph{shape correspondences} between sketches and photos. As \emph{shape} itself becomes almost unknown for unseen categories, discovering fine-grained correspondence becomes even harder.

\vspace{-0.4cm}
\section{Experiments}\label{sec:experiments}
\vspace{-0.3cm}

\noindent \textbf{Datasets:} For category-level adaptation, we use the Sketchy dataset \cite{sangkloy2016sketchy} which contains $125$ categories with $100$ photos each. Each photo has at least $5$ sketches with fine-grained associations. In contrast, QMUL-Shoe-V2 dataset \cite{bhunia2020sketch, pang2020solving, song2017deep} contains sketches of only one category (shoes) annotated with user ID and fine-grained sketch-photo correspondence, making it the only option for user/style-level adaptation. We consider users having at least $10$ sketch samples, which leads to a total of $306$ users having $5480$ sketches and corresponding $1877$ photos.

\noindent \textbf{Experimental Setup:} We demonstrate the potential of our framework in two scenarios.   \textbf{(a) Category-level adaptation:} Following \cite{yelamarthi2018zero, pang2019generalising}, we split the 125 Sketchy categories to 104 for training and the rest 21 for testing, ensuring no test categories are present in the 1000 ImageNet classes \cite{russakovsky2015imagenet}. We create random adaptation sets of $10$ photos each from each unseen category along with their respective sketches, leaving the rest photos and sketches for testing.  
\textbf{(b) User-level adaptation:}  We consider $60$ users with a sketch/gallery-photo size of $560/200$ for testing, and the rest \cut{remaining sketch-photos} for training. This ensures gallery-photos and users to be mutually exclusive for training and testing.  

\noindent \textbf{Implementation Details:} Inception-V3 network pretrained on ImageNet \cite{russakovsky2015imagenet} is used as our backbone feature extractor. The intermediate latent space $\mathbf{F}(\cdot)$ is of size $c =2048$, and we set $d=64$ as the dimension of our final joint-feature embedding layer $\mathbf{M}(\cdot)$. Following the  traditional supervised learning protocol \cite{song2017fine, bhunia2020sketch}, the Adam-optimiser \cite{kingma2014adam} with a learning rate of $0.0001$ is first used to pre-train the baseline model for $60$ epochs with a triplet loss, having a fixed margin of $0.3$ with batch size $16$. 
Thereafter, we add regularizer heads and perform meta-optimisation (Eqn. \ref{meta-opt}) for $40$ epochs. The reported performance uses only one inner-loop update during inference unless otherwise mentioned (ablative study done later). We use meta-batch size of $B=8$, and set the size of support and validation set as $K=5$. We use Adam as meta-optimiser with an outer-loop learning rate of $0.0001$. Note that the margin value of inner-loop triplet loss is meta-learned, that for outer-loop is set to $0.3$. Furthermore, we set  $\lambda$, $d'$, and $\mu'$ to $0.5$, $64$ and $0.2$, respectively. We implemented our framework in PyTorch \cite{paszke2017automatic} conducting experiments on a 11 GB Nvidia RTX 2080-Ti GPU. \doublecheck{We use pre-computed word-embeddings provided by Doodle2search [7], which ensures no leakage of class information.} {Please note that semantic relatedness module is used only for category-level adaptation on Sketchy; not for user style adaptation (on Shoe-V2), as no varying semantic concepts exist across different users. Please refer to $\S$ \textbf{Supplementary} (Supp.) for more details.}

\noindent \textbf{Evaluation Setup:} During inference, $\mathrm{K}$ sketch-photo pairs are used to construct triplets for adaptation, where negative images are randomly sampled.  We consider $\mathrm{k} \in \{1, 5, 10\}$ on Sketchy, and $\mathrm{k} \in \{1, 5\}$ on Shoe-V2 due to data constraint. For Shoe-V2, adaptation set is randomly sampled from each unseen user, and evaluation of adapted model is done on the rest samples. Only the final feature-embedding layer $\mathbf{M}$ is updated via inner loop update (Eqn.~\ref{innerloop}) for adaptation. For fair evaluation, we make sure the adaptation  and evaluation sets remain the same for all experiments. We evaluate the fine-grained retrieval performance using $\mathrm{Acc.@q}$ accuracy, i.e., percentage of sketches having true-match photos appearing in the top-q list.  Average accuracy is reported by repeating every experiment five times.

\vspace{-0.4cm}    
\subsection{Competitors}
\vspace{-0.2cm}    
To the best of our knowledge, there has been no prior work dealing with either category or user-level adaptation for SBIR. We thus design several baselines from \emph{four} different perspectives to justify our framework. \textbf{ (i) SOTA FG-SBIR Methods:} We compare with popular \emph{Triplet-SN} \cite{yu2016sketch} (Sketch-A-Net+ triplet loss) and  \emph{Triplet-HOLEF} \cite{song2017deep}. Results are cited at sketch-completion point for \emph{Triplet-RL}. Furthermore, we compare with \emph{Mixed-Jigsaw} employing self-supervised pre-training, and recently introduced StyleMeUP \cite{sain2021stylemeup}.
\textbf{(ii) Generalisation Approach:} \emph{CC-DG}  \cite{pang2019generalising}  aims to model a universal manifold of prototypical visual sketch traits that dynamically embeds sketch and photo, to generalise for unseen categories. Following a very recent few-shot classification work, we employ sequential distillation upon our baseline FG-SBIR model using $l_2$ loss on the absolute sketch/photo feature and evaluate it on unseen category/user without updating the model during inference. We term this few-shot competitor from non-MAML family as \emph{Distill}  \cite{tian2020rethinking}. \textbf{ (iii) Zero-Shot SBIR:} We also compare with four state-of-the-art ZS-SBIR methods, namely \emph{CVAE-Regress} \cite{yelamarthi2018zero}, \emph{Sem-Pyc} \cite{dutta2019semantically}, \emph{Doodle2Search} \cite{dey2019doodle}, \emph{SAKE} \cite{liu2019semantic}.\textbf{ (iv) Adaptation Based Approach:}  (a) We compare with standard \emph{Fine-Tuning} approach; (b)  Off-the-shelf \emph{MAML} \cite{finn2017MAML} has been employed on the top of our baseline FG-SBIR model additionally following the tricks introduced in \cite{antoniou2018trainMAML}. (c) Following \cut{the proposition of} \emph{ANIL} \cite{raghu2019rapid}, which only updates the final classification layer for few-shot classification, we  update final embedding layer $\textbf{M}$ within the inner loop in meta-optimisation process. We use fixed margin-hyperparameter value of $0.3$ for both inner and outer loop in case of MAML, sign-MAML \cite{fan2021sign} (recently introduced low-cost variant) and ANIL baselines.  Uniform backbone is used in all self-designed baselines and margin is meta-learned only in our final model.

Through preliminary experiments, we infer that adding a classification head is necessary for reasonable performance when dealing with multi-category FG-SBIR on Sketchy dataset. We thus add a classification head upon $\mathbf{F(\cdot)}$ for \emph{all our self-designed competitors (having uniform feature extractor) while experimenting on Sketchy}, and train using both triplet and classification losses with weights $1$ and $0.01$ respectively for a fair comparison.  

\vspace{-0.9cm}
\begin{table}[!hbt]
    \centering
        \caption{Comparing among our baseline FG-SBIR, naive Fine-tuning, Generalisation \cite{pang2019generalising} approach, and our proposed Category (Sketchy) and User(Shoe-V2)-adaptive FG-SBIR. $\mathrm{GAP_B}$ and $\mathrm{GAP_G}$ represent the Acc@1 gap of ours with Baseline and Generalisation respectively.}
    \footnotesize
    \resizebox{\columnwidth}{!}{%
    \begin{tabular}{ccc|cc|cc|cccc}
        \hline \hline
        \multirow{2}{*}{Datasets} & \multicolumn{2}{c}{Baseline} &
        \multicolumn{2}{|c}{Fine-Tuning} &
        \multicolumn{2}{|c}{Generalisation \cite{pang2019generalising}} & \multicolumn{4}{|c}{\textbf{Proposed} (k=5)} \\
        \cline{2-11}
         & Acc@1 & Acc@5 & Acc@1 & Acc@5 & Acc@1 & Acc@5 & Acc@1 & Acc@5 & $\mathrm{GAP_{B}}$ & $\mathrm{GAP_{G}}$ \\
        \hline 
        Sketchy (Category Level) & \blue{18.4\%} & 37.3\% & 18.5\% & 37.5\% & \blue{22.7\%} & 42.1\% & \red{28.1\%} & 51.8\% & \red{9.7$\uparrow$} & \red{5.4$\uparrow$} \\[2pt]
        Shoe-V2 (User Level) & \blue{33.7\%} & 70.2\% & 33.8\% & 70.2\% & \blue{33.8\%} & 70.4\% & \red{38.3\%} & 76.6\% & \red{4.6$\uparrow$} & \red{4.5$\uparrow$} \\
        \hline
    \end{tabular}
    }
    \label{tab:my_label1}
\vspace{-0.9cm}    
\end{table}

\vspace{-0.9cm}
\begin{table}
    \centering
    \caption{Performance analysis using different approaches.}
    \vspace{-0.05cm}
    \footnotesize
    \resizebox{\columnwidth}{!}{%
    \begin{tabular}{ccccccc|ccccccc}
        \hline
         &   &   & \multicolumn{2}{c}{Sketchy (Category)} & \multicolumn{2}{c|}{Shoe-V2 (User )} & & &   & \multicolumn{2}{c}{Sketchy (Category)} & \multicolumn{2}{c}{Shoe-V2 (User )} \\
        \cline{4-7}\cline{10-14}
         & & & Acc@1 & Acc@5 & Acc@1 & Acc@5 & & & & Acc@1 & Acc@5 & Acc@1 & Acc@5\\
        \hline
        & \multicolumn{2}{c}{Our Baseline } & {18.4\%} & {37.3\%} & {33.7\%} & {70.2\%} & \multirow{12}{*}{\rotatebox[origin=c]{90}{Adaptation Based Approaches}} 
        & \multirow{3}{*}{Fine-Tuning} & k=1 & 18.4\% & 37.3\% & 33.7\% & 70.2\% \\
        & \multicolumn{2}{c}{Our Baseline + Reg. } & {19.2\%} & {39.6\%} & {33.9\%} & {71.3\%} & & & k=5 & 18.5\% & 37.5\% & 33.8\% & 70.2\% \\
          & \multicolumn{2}{c}{Upper-Bound} & 29.8\% & 53.7\% & -- & -- & & & k=10 & 18.6\% & 37.5\% & -- & -- \\
        \cdashline{1-7}\cdashline{9-14}
        \multirow{4}{*}{\rotatebox[origin=c]{90}{SOTA}} 
         & \multicolumn{2}{c}{Triplet-SN \cite{yu2016sketchAnet}} & 15.3 \% & 34.0\% & 28.5\% & 67.3\%  & & \multirow{3}{*}{MAML \cite{finn2017MAML}} & k=1 & 19.5\% & 38.7\% & 34.2\% & 70.7\% \\ 
         & \multicolumn{2}{c}{Triplet-HOLEF \cite{song2017deep}} & 16.7\% & 35.9\% & 31.4\% & 69.1\% & & & k=5 & 22.8\% & 42.3\% & 35.5\% & 74.6\% \\
        & \multicolumn{2}{c}{Triplet-RL  \cite{bhunia2020sketch}} & 4.7\% & 7.8\% & 34.1\% & 70.2\% & & & k=10 & 26.4\% & 48.9\% & -- & -- \\\cdashline{9-14}
         & \multicolumn{2}{c}{Mixed-Jigsaw \cite{pang2019generalising}} & 16.7\% & 34.3\% & 33.5\% & 71.4\% & & \multirow{2}{*}{sign-MAML \cite{fan2021sign}} & k=1 & 19.1\% & 38.2\% & 33.8\%  & 69.6\%  \\   
        & \multicolumn{2}{c}{StyleMeUp \cite{sain2021stylemeup}} & 19.6\% & 39.7\% & 36.4\% & 81.8\%  & &  & k=5 & 20.5\% & 39.6\% & 34.1\% & 70.8\% \\ \cdashline{9-14}
         \cdashline{1-7}
        \multirow{2}{*}{\rotatebox[origin=c]{90}{GA}} 
         & \multicolumn{2}{c}{CC-DG \cite{pang2019generalising}} & 22.7\% & 42.1\% & 33.8\% & 70.4\% & & \multirow{3}{*}{ANIL \cite{raghu2019rapid}} & k=1 & 19.7\% & 38.9\% & 34.5\% & 70.9\% \\
         & \multicolumn{2}{c}{Distill(non-MAML)\cite{pang2019generalising}} & {18.9\%} & {38.1\%} & {33.9\%} & {70.9\%} & & & k=5 & 23.2\% & 42.8\% & 35.7\% & 75.3\% \\ \cdashline{1-7}
        \multirow{4}{*}{\rotatebox[origin=c]{90}{ZS-SBIR}} 
         & \multicolumn{2}{c}{CVAE-Regress \cite{yelamarthi2018zero}} & 2.4\% & 9.5\% & 1.8\% & 3.1\% & & & k=10 & 26.9\% & 48.3\% & -- & -- \\ \cdashline{9-14}
         & \multicolumn{2}{c}{Sem-Pyc \cite{dutta2019semantically}} & 4.9\% & 17.3\% & 2.1\% & 4.7\% & & \multirow{4}{*}{\textbf{Ours}} & k=1 & 21.8\% & 42.5\% & 34.9\% & 71.4\% \\
         & \multicolumn{2}{c}{Doodle2Search \cite{dey2019doodle}} & 14.8\% & 34.5\% & 28.1\% & 66.9\% & & & k=5 & 28.1\% & 51.8\% & 38.3\% & 76.6\% \\
          & \multicolumn{2}{c}{SAKE \cite{liu2019semantic}} & 6.4\% & 20.3\% & 3.6\% & 5.7\% & & & k=10 & 32.7\% & 53.5\% & -- & -- \\\hline
    \end{tabular}
    }
    \label{tab:my_label2}
    \vspace{-0.6cm}
\end{table}

\vspace{-0.5cm}
\subsection{Performance Analysis}\label{sec:category-perform}
\begin{figure}[!hbt]
 \centering
 \includegraphics[width=\linewidth]{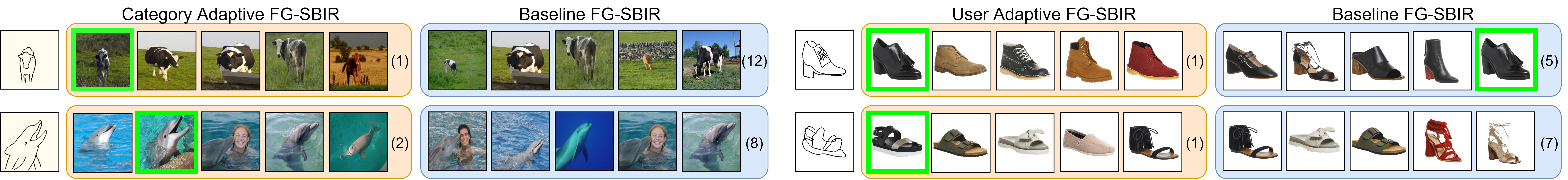}
\vspace{-0.7cm}
\caption{Category (left) and user (right) level adaptive model vs. baseline [($\cdot$): matching photo's rank] (more in supplementary). } 
\label{retrieval_qual}
\vspace{-0.8cm}
\end{figure}

\vspace{-0.1cm}
Table \ref{tab:my_label1} shows our adaptation (5-shot) based framework to outperform baseline FG-SBIR ($\S$ \ref{basemodel}) and Generalisation based approach \cite{pang2019generalising} by a significant margin of $9.7\%$ ($4.6\%$) and $5.4\%$ ($4.5\%$) in Acc@1, respectively for category(user) level adaptation. Furthermore, we compare with \emph{four} different classes of alternative approaches in Table \ref{tab:my_label2}, including an upper-bound for Sketchy, where we re-train the model on the testing (unseen) categories with available samples. 
\noindent \textbf{Category-Level Adaptation:} We can make the following observations: \emph{(i) SOTA FG-SBIR Methods:} Almost every existing state-of-the-arts model including our baseline performs poorly on unseen testing sketchy classes, indicating that adaptation is necessary. Note that, among them our baseline FG-SBIR model is notably better compared to earlier SOTA triplet-loss based frameworks, \emph{Triplet-SN} and \emph{Triplet-HOLEF}, due to  more recent backbone feature extractor (Inception-V3) with spatial attention.  While \emph{Triplet-RL} fails to converge for large Sketchy dataset as the reward diminishes to zero during RL-based fine-tuning, \emph{Mixed-Jigsaw}/\emph{StyleMeUP} are found to be less effective \cite{pang2020solving} on Sketchy. \emph{(ii) Generalisation Approach:} \emph{CC-DG} (our re-implementation) is the only exception that performs comparatively better than other SOTA methods, as it models category agnostic abstract sketch traits \cite{pang2019generalising} for better cross-category generalisation. Nevertheless, it does not provide any option to obtain category specialised model during inference; hence its performance is much lower than our adaptation based pipeline. \emph{Distill} (non-MAML baseline) gives very marginal gain over our baseline.
\emph{(iii) Zero-Shot SBIR:} Every ZS-SBIR method was designed for category-level retrieval, not instance-level, thus limiting its efficacy in FG-SBIR.
\doublecheck{Doodle2search \cite{dey2019doodle} performs relatively better due to triplet loss (unlike the rest), which is critical for instance-level matching in FG-SBIR.}
\emph{(iv) Adaptation Based Approach:} Notably naive  \emph{Fine-Tuning} hardly helps over few-shot setting. It can be seen that \emph{ANIL} performs better than \emph{MAML}. This suggests that simplifying the inner loop update in MAML to reduce the high computational cost associated with second order gradients over a large parameter space is indeed useful. 
However, its performance is still lower than ours.
We also tried our first-order approximated version of MAML and very recently introduced sign-MAML \cite{fan2021sign}, but found no significant difference to MAML. To summarise, our meta-learning based \emph{adaptive fine-grained SBIR} framework outperforms existing SOTA methods, alternative generalisation and zero-shot approaches by a large margin, as well as exceeds some strong few-shot baselines by a significant margin. (v) \emph{Most importantly}, accuracy even after adding all the respective regularizers to baseline FG-SBIR model falls behind our method by a significant margin of \doublecheck{$9-10\%$ (Sketchy) and $4-5\%$ (Shoe) -- thus proving the contribution of our bi-level meta-learning  framework.} The  Qualitative results are shown in Figure \ref{retrieval_qual}.\\
\noindent \textbf{User-Level Adaptation:} Compared to the striking boost obtained in the category-level adaptation experiments by our method, improvements for user level adaptation (Table \ref{tab:my_label1} and \ref{tab:my_label2}) are relatively small (difference of $4.6\%$  $\mathrm{Acc@1}$) compared to our baselines.  One explanation is that modelling user-specific subtle differences is more challenging compared to category-level modelling. Nevertheless, the overall pattern   is fairly similar  to that of category-level adaptation and a same set of conclusions can be drawn regarding the effectiveness of our approach.


\vspace{-.5cm}

\vspace{-0.5cm}  
\subsection{Ablative Studies}\label{ablation}
\vspace{-.3cm}  
\noindent \textbf{Contributions of Regularizers:} Ablative studies in Table \ref{tab:abalation} evaluate the contribution of different regularizers used for optimisation. (i) We notice $\mathrm{L_D}$, mitigating the domain gap between sketch and photos, has a relatively uniform effect
\begin{wraptable}[7]{l}{0.5\linewidth}
\scriptsize
\centering
    \vspace{-0.6cm}
    \caption{Ablative study ($\mathrm{Acc@1})$; $k = 5$}  
    \vspace{-0.3cm}
    \begin{tabular}{cccc|ccccc}
        \hline
        \multirow{2}{*}{$L_D$} & \multirow{2}{*}{$L_S$} & \multirow{2}{*}{$L_C$} &  Sketchy &  \multirow{2}{*}{$L_D$} & \multirow{2}{*}{$L_{ud}$} & Shoe-V2 \\
         & & & Category Level & & & User Level \\
        \hline
        \checkmark & \checkmark & \checkmark & 28.1\% & \checkmark & \checkmark & 38.3\% \\
        $\times$ & \checkmark & \checkmark & 26.3\% & $\times$ & \checkmark & 37.1\% \\
        $\times$ & $\times$ & \checkmark & 23.7\% & $\times$ & $\times$ & 35.8\% \\
        $\times$ & $\times$ & $\times$ & 16.5\% & - & - & - \\
        \hline
    \end{tabular}
    \label{tab:abalation}
\end{wraptable} 
on both category and user level adaptation.
(ii) Classification head, employing a classification loss $\mathrm{L_C}$ is the most critical one while dealing with multi-category FG-SBIR on Sketchy dataset to maintain class discriminative information. Removing only $\mathrm{L_C}$ leads to a drop of $7.5\%$ under $\mathrm{k=5}$ on Sketchy. For multi-category FG-SBIR, class specific grouping (classification loss) followed by instance specific separation (hard triplet loss) is necessary. (iii) Semantic loss $\mathrm{L_S}$ plays a vital role in adapting to unseen Sketchy categories \cut{by preserving semantic relatedness in the intermediate latent space} to transfer knowledge from seen training classes to unseen ones. (iv) Removing $\mathrm{L_{ud}}$, that helps in learning discriminative information across different users' sketching styles, drops accuracy by $1.8\%$ for user-level adaptation $\mathrm{k=5}$ on Shoe-V2 dataset.

\vspace{-0.5cm}
\begin{figure}
\centering
\begin{minipage}{.47\textwidth}
  \centering
  \includegraphics[width=.9\linewidth]{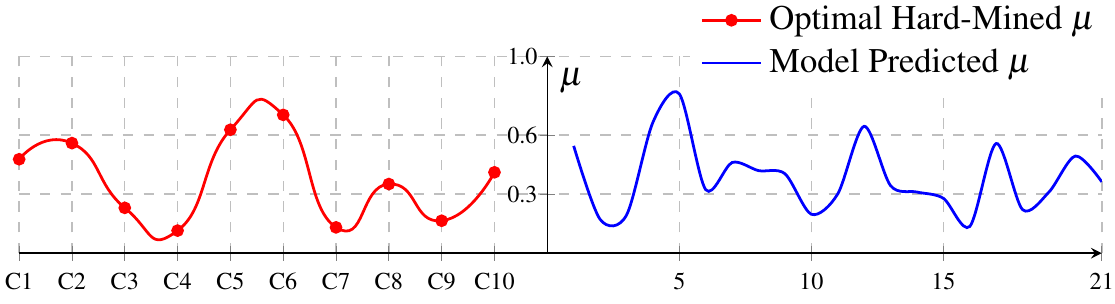}
  \vspace{-0.25cm}
  \captionof{figure}{(a) Hard-mined $\mu$ over 10 random classes. \cut{(bat, cabin, cow, dolphin, door, giraffe, helicopter, mouse, pear, raccoon).} (b) Model predicted $\mu$ over 21 testing classes. (both Sketchy)}
  \label{fig:ablation1}
\end{minipage}\hspace{0.2cm}
\begin{minipage}{.47\textwidth}
  \centering
  \vspace{-0.2cm}
  \includegraphics[width=\linewidth]{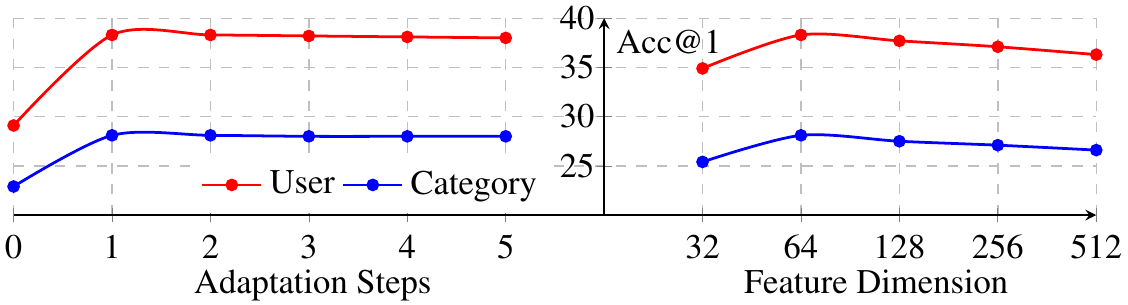}
  \vspace{-0.4cm}
  \captionof{figure}{Varying (a) adaptation steps (b) feature dimension (k=5).}
  \label{fig:ablation2}
\end{minipage}
\end{figure}
\vspace{-0.3cm}

 

\vspace{-0.5cm}
\keypoint{{Effect of meta-learning $\mathbf{\mu}$:}} A direct way of judging the contribution of the learnable margin hyper-parameter $\mu$, is to replace it by a fixed value of $0.3$ (optimised) in the inner loop loss calculation. Consequently, we notice a significant drop of  $1.8\%$ $\mathrm{Acc@1}$  ($\mathrm{k=5}$) for category-level adaptation respectively.  To verify if $\mu$ really varies across different Sketchy classes, we randomly choose 10 classes to perform exhaustive hyper-parameter search with bin size $0.05$ around global optimum $\mu$ of $0.3$. Figure \ref{fig:ablation1} (a) shows that the optimal value indeed varies. Such a search for the test categories obviously is infeasible due to the lack of data. In Figure \ref{fig:ablation1} (b), our model-predicted average $\mu$ over different classes is plotted against the $21$ Sketchy testing classes -- the value clearly varies contributing to the performance boost.   However, the effect of learnable margin is almost negligible (a $0.06\%$ boost) in case of user-level adaptation as all images belong to a single category for Shoe-V2 dataset. The reported numbers on Shoe-V2 are hence based on fixed inner loop margin value of $0.3$.

%
\vspace{-0.1cm} 
\keypoint{Cross-Dataset Adaptation:}  Model trained on Sketchy training classes gives $10.3\%$ Acc@1 on QMUL-ShoeV2 \cite{bhunia2020sketch}, however upon adaptation using 5 (10) random sketch-photo pairs from respective datasets, Acc@1 jumps to $22.3\%$ ($26.4\%$), respectively. In contrast to complete-training dataset supervised performance of $33.7\%$, this demonstrates our \cut{fairly significant} \emph{cross-dataset} generalisation capability.  

 
\keypoint{{Further Analysis:}}  (i) From  Figure \ref{fig:ablation2} (a), the optimal accuracy is observed at joint feature-embedding space dimension d = 64. (ii) The number of gradient-update steps are varied as well -- Figure \ref{fig:ablation2} (b) shows that a single gradient step update, used in all the experiments, provides the highest performance gain. (iii) We explore other word-embedding techniques for auxiliary semantic loss $L_S$, but found that Glove \cite{pennington2014glove} and word2vec \cite{mikolov2013efficient} give  $27.5\%$ and $27.7\%$ for unseen category-level adaptation compared to $28.1\%$ in case of fast-text \cite{bojanowski2017enriching} using $k=5$. (iv) One could have used a simple sum or average pooling operation to accumulate information for predicting learnable $\mu$. However a performance drop ($\mathrm{Acc@1}$) of $1.2\%/0.6\%/0.8\%$ during sum/average/max pooling operation under category setup ($\mathrm{k=5}$) demonstrates the relevance of relational network which encodes the joint relationship between all sketch-photo pairs in support set for predicting learnable $\mu$. (v) For 5-shot single gradient step adaptation on our baseline model, an Intel(R) Xeon(R) W-2123 CPU @ 3.60GHz takes $32.1$ms. 

\vspace{-0.4cm}
\begin{figure}
\begin{wrapfigure}[6]{l}{0.6\linewidth}
\vspace{-1cm}
    \includegraphics[width=0.49\linewidth]{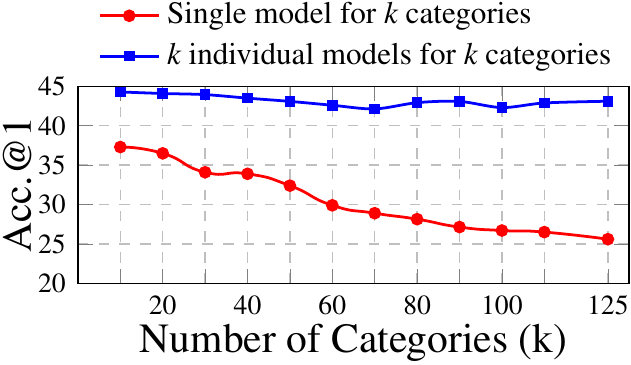}  
    \includegraphics[width=0.49\linewidth]{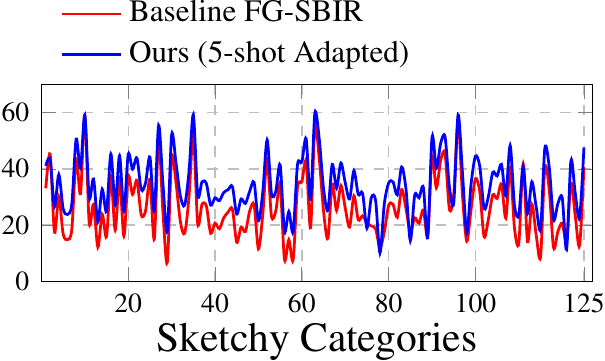}
\vspace{-1.2cm}
\end{wrapfigure}
Fig. 6: Graph (left) shows how {baseline} FG-SBIR model falls behind with rise in categories (also during training) to be served by a single FG-SBIR model, compared to having {multiple individual models} for each category in Sketchy. Right shows benefit of adaptation.
\vspace{-1cm} 
\label{fig:seen}
\end{figure}
\setcounter{figure}{6}


\keypoint{\emph{Is adaptation useful even for seen classes?:}}  Instead of designing individual FG-SBIR model for each category, cost-effective deployment requires a single model handling instance-specific retrieval from multiple categories. However, as the number of categories to be handled by a single model increases, the retrieval performance starts decreasing drastically even for the seen classes that the model has been trained upon, as shown in Figure 6. For instance, single model trained from all the $125$ Sketchy classes, gives average Acc@1 of only $25.6\%$ on the test set. Whereas, on re-training $125$ individual models for each category, the same value rises up to $43.1\%$, although it is quite impractical to have 125 separate models. In such a scenario, using our method on a single model to adapt to just one of the \text{seen} training categories with only 5/10 sketch-photo pairs, we obtain an average Acc@1 of $34.9/38.7\%$. 

\vspace{-0.5cm}
\section{Conclusion}
\vspace{-0.3cm}
We have introduced a FG-SBIR framework which retains a single model that can quickly adapt to (i) sketching style of a particular user, or (ii) a new category, with just very few examples during the inference process. To this end, we design a meta-learning framework based on the existing MAML model but with crucial modification to our retrieval problem, including a simplified inner loop optimisation and introduction of the learnable contrastive loss margin to the meta-learning process. The intermediate latent space, upon which meta-learning is performed, is further constrained using three additional regularisation losses to facilitate learning the adaptation process during meta-optimisation.

\clearpage
%
%
\bibliographystyle{splncs04}
\bibliography{egbib}

\begin{thebibliography}{10}
\providecommand{\url}[1]{\texttt{#1}}
\providecommand{\urlprefix}{URL }
\providecommand{\doi}[1]{https://doi.org/#1}

\bibitem{antoniou2018trainMAML}
Antoniou, A., Edwards, H., Storkey, A.: How to train your maml. In: ICLR (2018)

\bibitem{bhunia2021more}
Bhunia, A.K., Chowdhury, P.N., Sain, A., Yang, Y., Xiang, T., Song, Y.Z.: More
  photos are all you need: Semi-supervised learning for fine-grained sketch
  based image retrieval. In: CVPR (2021)

\bibitem{bhunia2021vectorization}
Bhunia, A.K., Chowdhury, P.N., Yang, Y., Hospedales, T.M., Xiang, T., Song,
  Y.Z.: Vectorization and rasterization: Self-supervised learning for sketch
  and handwriting. In: CVPR (2021)

\bibitem{bhunia2022diy}
Bhunia, A.K., Gajjala, V.R., Koley, S., Kundu, R., Sain, A., Xiang, T., Song,
  Y.Z.: Doodle it yourself: Class incremental learning by drawing a few
  sketches. In: CVPR (2022)

\bibitem{bhunia2021metahtr}
Bhunia, A.K., Ghose, S., Kumar, A., Chowdhury, P.N., Sain, A., Song, Y.Z.:
  Metahtr: Towards writer-adaptive handwritten text recognition. In: CVPR
  (2021)

\bibitem{bhunia2022subset}
Bhunia, A.K., Koley, S., Khilji, A.F.U.R., Sain, A., Chowdhury, P.N., Xiang,
  T., Song, Y.Z.: Sketching without worrying: Noise-tolerant sketch-based image
  retrieval. In: CVPR (2022)

\bibitem{bhunia2020sketch}
Bhunia, A.K., Yang, Y., Hospedales, T.M., Xiang, T., Song, Y.Z.: Sketch less
  for more: On-the-fly fine-grained sketch based image retrieval. In: CVPR
  (2020)

\bibitem{bojanowski2017enriching}
Bojanowski, P., Grave, E., Joulin, A., Mikolov, T.: Enriching word vectors with
  subword information. TACL  (2017)

\bibitem{choi2020adaptiveScene}
Choi, M., Choi, J., Baik, S., Kim, T.H., Lee, K.M.: Scene-adaptive video frame
  interpolation via meta-learning. In: CVPR (2020)

\bibitem{pinaki2022PartialSBIR}
Chowdhury, P.N., Bhunia, A.K., Gajjala, V.R., Sain, A., Xiang, T., Song, Y.Z.:
  Partially does it: Towards scene-level fg-sbir with partial input. In: CVPR
  (2022)

\bibitem{chowdhury2022fs}
Chowdhury, P.N., Sain, A., Bhunia, A.K., Xiang, T., Gryaditskaya, Y., Song,
  Y.Z.: Fs-coco: Towards understanding of freehand sketches of common objects
  in context. In: ECCV (2022)

\bibitem{collomosse2019livesketch}
Collomosse, J., Bui, T., Jin, H.: Livesketch: Query perturbations for guided
  sketch-based visual search. In: CVPR (2019)

\bibitem{collomosse2017sketching}
Collomosse, J., Bui, T., Wilber, M.J., Fang, C., Jin, H.: Sketching with style:
  Visual search with sketches and aesthetic context. In: ICCV (2017)

\bibitem{dey2019doodle}
Dey, S., Riba, P., Dutta, A., Llados, J., Song, Y.Z.: Doodle to search:
  Practical zero-shot sketch-based image retrieval. In: CVPR (2019)

\bibitem{dou2019domain}
Dou, Q., de~Castro, D.C., Kamnitsas, K., Glocker, B.: Domain generalization via
  model-agnostic learning of semantic features. In: NeurIPS (2019)

\bibitem{dutta2019semantically}
Dutta, A., Akata, Z.: Semantically tied paired cycle consistency for zero-shot
  sketch-based image retrieval. In: CVPR (2019)

\bibitem{dutta2020semantically}
Dutta, A., Akata, Z.: Semantically tied paired cycle consistency for any-shot
  sketch-based image retrieval. IJCV  (2020)

\bibitem{SBIR_imbalance}
Dutta, T., Singh, A., Biswas, S.: Adaptive margin diversity regularizer for
  handling data imbalance in zero-shot sbir. In: ECCV (2020)

\bibitem{fan2021sign}
Fan, C., Ram, P., Liu, S.: Sign-maml: Efficient model-agnostic meta-learning by
  signsgd. arXiv preprint arXiv:2109.07497  (2021)

\bibitem{finn2017MAML}
Finn, C., Abbeel, P., Levine, S.: Model-agnostic meta-learning for fast
  adaptation of deep networks. In: ICML (2017)

\bibitem{finn2018probabilistic}
Finn, C., Xu, K., Levine, S.: Probabilistic model-agnostic meta-learning. In:
  NeurIPS (2018)

\bibitem{fu2015zero}
Fu, Z., Xiang, T., Kodirov, E., Gong, S.: Zero-shot object recognition by
  semantic manifold distance. In: CVPR (2015)

\bibitem{ganin2015unsupervised}
Ganin, Y., Lempitsky, V.: Unsupervised domain adaptation by backpropagation.
  In: ICML (2015)

\bibitem{garcia2020adaptiveEmotion}
Garcia-Ceja, E., Riegler, M., Kvernberg, A.K., Torresen, J.: User-adaptive
  models for activity and emotion recognition using deep transfer learning and
  data augmentation. User Modeling and User-Adapted Interaction  (2020)

\bibitem{horiguchi2019significance}
Horiguchi, S., Ikami, D., Aizawa, K.: Significance of softmax-based features in
  comparison to distance metric learning-based features. IEEE-TPAMI  (2019)

\bibitem{tim2020metaSurvey}
Hospedales, T., Antoniou, A., Micaelli, P., Storkey, A.: Meta-learning in
  neural networks: A survey. arXiv preprint arXiv:2004.05439  (2020)

\bibitem{hsieh2015facial}
Hsieh, P.L., Ma, C., Yu, J., Li, H.: Unconstrained realtime facial performance
  capture. In: CVPR (2015)

\bibitem{kingma2014adam}
Kingma, D.P., Ba, J.: Adam: A method for stochastic optimization. arXiv
  preprint arXiv:1412.6980  (2014)

\bibitem{lane2011activity}
Lane, N.D., Xu, Y., Lu, H., Hu, S., Choudhury, T., Campbell, A.T., Zhao, F.:
  Enabling large-scale human activity inference on smartphones using community
  similarity networks (csn). In: UbiComp (2011)

\bibitem{li2014fine}
Li, Y., Hospedales, T.M., Song, Y.Z., Gong, S.: Fine-grained sketch-based image
  retrieval by matching deformable part models. In: BMVC (2014)

\bibitem{li2017metaSGD}
Li, Z., Zhou, F., Chen, F., Li, H.: Meta-sgd: Learning to learn quickly for
  few-shot learning. arXiv preprint arXiv:1707.09835  (2017)

\bibitem{liu2017deep}
Liu, L., Shen, F., Shen, Y., Liu, X., Shao, L.: Deep sketch hashing: Fast
  free-hand sketch-based image retrieval. In: CVPR (2017)

\bibitem{liu2019semantic}
Liu, Q., Xie, L., Wang, H., Yuille, A.: Semantic-aware knowledge preservation
  for zero-shot sketch-based image retrieval. In: ICCV (2019)

\bibitem{mikolov2013efficient}
Mikolov, T., Chen, K., Corrado, G., Dean, J.: Efficient estimation of word
  representations in vector space. In: ICLR (2014)

\bibitem{oreshkin2018tadam}
Oreshkin, B., L{\'o}pez, P.R., Lacoste, A.: Tadam: Task dependent adaptive
  metric for improved few-shot learning. In: NeurIPS (2018)

\bibitem{pang2019generalising}
Pang, K., Li, K., Yang, Y., Zhang, H., Hospedales, T.M., Xiang, T., Song, Y.Z.:
  Generalising fine-grained sketch-based image retrieval. In: CVPR (2019)

\bibitem{pang2017cross}
Pang, K., Song, Y.Z., Xiang, T., Hospedales, T.M.: Cross-domain generative
  learning for fine-grained sketch-based image retrieval. In: BMVC (2017)

\bibitem{pang2020solving}
Pang, K., Yang, Y., Hospedales, T.M., Xiang, T., Song, Y.Z.: Solving
  mixed-modal jigsaw puzzle for fine-grained sketch-based image retrieval. In:
  CVPR (2020)

\bibitem{paszke2017automatic}
Paszke, A., Gross, S., Chintala, S., Chanan, G., Yang, E., DeVito, Z., Lin, Z.,
  Desmaison, A., Antiga, L., Lerer, A.: Automatic differentiation in {PyTorch}.
  In: NeurIPS Autodiff Workshop (2017)

\bibitem{pennington2014glove}
Pennington, J., Socher, R., Manning, C.D.: Glove: Global vectors for word
  representation. In: EMNLP (2014)

\bibitem{raghu2019rapid}
Raghu, A., Raghu, M., Bengio, S., Vinyals, O.: Rapid learning or feature reuse?
  towards understanding the effectiveness of maml. In: ICLR (2020)

\bibitem{russakovsky2015imagenet}
Russakovsky, O., Deng, J., Su, H., Krause, J., Satheesh, S., Ma, S., Huang, Z.,
  Karpathy, A., Khosla, A., Bernstein, M., et~al.: Imagenet large scale visual
  recognition challenge. IJCV  (2015)

\bibitem{rusu2019LEO}
Rusu, A.A., Rao, D., Sygnowski, J., Vinyals, O., Pascanu, R., Osindero, S.,
  Hadsell, R.: Meta-learning with latent embedding optimization. In: ICLR
  (2019)

\bibitem{Sketch3T}
Sain, A., Bhunia, A.K., Potlapalli, V., Chowdhury, P.N., Xiang, T., Song, Y.Z.:
  Sketch3t: Test-time training for zero-shot sbir. In: CVPR (2022)

\bibitem{BMVC_hierarchy}
Sain, A., Bhunia, A.K., Yang, Y., Xiang, T., Song, Y.Z.: Cross-modal
  hierarchical modelling forfine-grained sketch based image retrieval. In: BMVC
  (2020)

\bibitem{sain2021stylemeup}
Sain, A., Bhunia, A.K., Yang, Y., Xiang, T., Song, Y.Z.: Stylemeup: Towards
  style-agnostic sketch-based image retrieval. In: CVPR (2021)

\bibitem{sampaio2020sketchformer}
Sampaio Ferraz~Ribeiro, L., Bui, T., Collomosse, J., Ponti, M.: Sketchformer:
  Transformer-based representation for sketched structure. In: CVPR (2020)

\bibitem{sangkloy2016sketchy}
Sangkloy, P., Burnell, N., Ham, C., Hays, J.: The sketchy database: learning to
  retrieve badly drawn bunnies. ACM TOG  (2016)

\bibitem{shen2018zero}
Shen, Y., Liu, L., Shen, F., Shao, L.: Zero-shot sketch-image hashing. In: CVPR
  (2018)

\bibitem{snell2017prototypical}
Snell, J., Swersky, K., Zemel, R.S.: Prototypical networks for few shot
  learning. In: NeurIPS (2017)

\bibitem{soh2017UI}
Soh, H., Sanner, S., White, M., Jamieson, G.: Deep sequential recommendation
  for personalized adaptive user interfaces. In: IUI (2017)

\bibitem{song2017fine}
Song, J., Song, Y.Z., Xiang, T., Hospedales, T.M.: Fine-grained image
  retrieval: the text/sketch input dilemma. In: BMVC (2017)

\bibitem{song2017deep}
Song, J., Yu, Q., Song, Y.Z., Xiang, T., Hospedales, T.M.: Deep
  spatial-semantic attention for fine-grained sketch-based image retrieval. In:
  ICCV (2017)

\bibitem{tian2020rethinking}
Tian, Y., Wang, Y., Krishnan, D., Tenenbaum, J.B., Isola, P.: Rethinking
  few-shot image classification: a good embedding is all you need? In: ECCV
  (2020)

\bibitem{wang2015sketch}
Wang, F., Kang, L., Li, Y.: Sketch-based 3d shape retrieval using convolutional
  neural networks. In: CVPR (2015)

\bibitem{weinberger2009metric_learn_margin}
Weinberger, K.Q., Saul, L.K.: Distance metric learning for large margin nearest
  neighbor classification. JMLR  (2009)

\bibitem{yelamarthi2018zero}
Yelamarthi, S.K., Reddy, S.K., Mishra, A., Mittal, A.: A zero-shot framework
  for sketch based image retrieval. In: ECCV (2018)

\bibitem{yu2016sketch}
Yu, Q., Liu, F., Song, Y.Z., Xiang, T., Hospedales, T.M., Loy, C.C.: Sketch me
  that shoe. In: CVPR (2016)

\bibitem{QianIJCV2020}
Yu, Q., Song, J., Song, Y.Z., Xiang, T., Hospedales, T.M.: Fine-grained
  instance-level sketch-based image retrieval. IJCV  (2020)

\bibitem{yu2016sketchAnet}
Yu, Q., Yang, Y., Liu, F., Song, Y.Z., Xiang, T., Hospedales, T.M.:
  Sketch-a-net: A deep neural network that beats humans. IJCV  (2017)

\bibitem{zhang2018generative}
Zhang, J., Shen, F., Liu, L., Zhu, F., Yu, M., Shao, L., Tao~Shen, H.,
  Van~Gool, L.: Generative domain-migration hashing for sketch-to-image
  retrieval. In: ECCV (2018)

\end{thebibliography}
\end{document}